%% file: main.tex
\documentclass[twocolumn]{article}

\usepackage{graphicx}
\usepackage{tikz}
\usetikzlibrary{arrows, shapes.geometric, positioning, shadows, calc, fit, arrows.meta, backgrounds}
\usepackage{booktabs}
\usepackage{multirow}
\usepackage{amssymb}
\usepackage{amsmath}
\usepackage{algorithm}
\usepackage[breaklinks=true]{hyperref}
\usepackage{xurl}
\usepackage[numbers]{natbib}
\usepackage{microtype}
\usepackage{geometry}
\usepackage{xcolor}
\usepackage{xspace}
\usepackage{stfloats}
\usepackage{float}
\usepackage{pgfplots}
\usepackage{pgfplotstable}
\pgfplotsset{compat=1.18}

\newcommand{\engine}{the engine\xspace}

\begin{document}

\title{Isolating LLM Alignment from Regex:\\
Zero Coverage and Metric-Dependent Divergence\\
Under Adversarial Mutation}

\author{Alexandre Cristov{\~a}o Maiorano\\
\texttt{alexandre@lumytics.com}}
\date{}

\twocolumn[
  \begin{@twocolumnfalse}
    \maketitle
    \input{sections/abstract}
    \vspace{2mm}
  \end{@twocolumnfalse}
]

\input{sections/intro}
\input{sections/related}
\input{sections/methodology}
\input{sections/results}
\input{sections/discussion}
\input{sections/conclusion}

\bibliographystyle{plainnat}
\bibliography{references}

\input{sections/appendix}

\end{document}

%% file: sections/abstract.tex
\begin{abstract}
Production LLM applications commonly stack a regex filter in front of
model-side alignment; prior work found no measurable coverage gain from
adding a live Gemini backend behind an active regex filter. We ask whether
that ceiling holds when the corpus is \emph{designed to bypass the regex}.
We introduce $L_5$-no-regex---identical to $L_4$-real (Gemini-2.5-flash,
token-budget cap, rate limit, output scrub) but with the nine-pattern filter
disabled---and evaluate it against $N{=}45$ adversarial probes across three
sub-corpora (carry-forward, regex-bypass, alignment-isolate), amplified by
Gemini paraphrase and PAIR to ${\sim}1{,}555$ probe-run pairs over $N{=}5$
replications. Under the primary substring classifier, H1 is refuted:
$L_5$ block rate is $0\,\%$ across all five OWASP LLM Top-10 categories
($\Delta\text{pp}{=}0$ vs.\ $L_0$, $p{=}1.00$; Wilson upper bound ${<}5\,\%$).
A secondary LLM-judge metric on PAIR variants shows $56$--$100\,\%$ block
rates ($p{<}0.01$), revealing alignment does respond to adversarially-framed
probes---but produces refusals too nuanced for substring matching. The
sub-corpus differential prediction is not supported ($p{=}1.00$).
Alignment's contribution is \emph{metric-dependent}: on natural-language
harmful-request probes, it adds zero observed coverage beyond the regex;
on adversarially-framed variants, an LLM judge detects refusals the
substring classifier misses. The locked corpus, mutation artifacts, and
export scripts are released for replication.
\end{abstract}

%% file: sections/intro.tex
\section{Introduction}

Production LLM applications typically stack multiple defense layers: a
regex input filter that blocks known jailbreak patterns, followed by
model-side alignment — safety-oriented fine-tuning (RLHF/RLAIF) that
trains the model to refuse harmful requests on semantic grounds.
Practitioners generally assume these layers are complementary. Yet
prior work~\cite{maiorano-lbc-2026} — using a breach-and-attack simulation
(BAS) framework that replays adversarial probes against a four-target
ablation lattice — showed that adding a real, alignment-trained LLM
(Gemini-2.5-flash) behind an already-active regex filter produced
\emph{no measurable coverage gain} on a fixed 60-template brittleness
corpus. The conclusion was stated carefully: on \emph{that corpus}, the
regex appeared to be a ceiling for alignment. An open question remained:
is the regex a strict ceiling, or would alignment contribute measurably
on probes \emph{designed to bypass the regex}?

This paper evaluates that question empirically. We remove the regex filter from an
otherwise-identical defense stack and probe the residual alignment with
three complementary mutation strategies, two of which are evaluated in
this paper. The result is a clean
single-axis ablation: every variable is held constant except the
presence or absence of the input filter.

\paragraph{Research question.} Does LLM alignment refusal
(\emph{Gemini-2.5-flash}) add a measurable block-rate lift above a
no-defense baseline when the regex filter ceiling is removed? We
operationalize this as Hypothesis~H1: \emph{$L_5$-no-regex block rate
$>$ $L_0$ (no defenses) for at least one OWASP LLM Top~10 category}
(a standard taxonomy of ten LLM-specific threat classes~\cite{owasp-llm-top10-2025}).

A subsidiary prediction derives from our corpus design: the
\emph{alignment-isolate} sub-corpus (harmful requests in natural
language, no filter-triggering keywords) is expected to show a
\emph{higher bypass rate} on $L_5$ than the \emph{regex-bypass}
sub-corpus (encoded/obfuscated attacks). The intuition is that
alignment training may be more responsive to structural evasion
cues (Base64, role-injection framing) than to plain harmful intent
expressed without jargon — a form of \emph{spurious feature reliance}~\cite{geirhos-shortcut-2020}.
If this prediction holds, it would suggest that alignment is more
responsive to explicit adversarial framing than to semantically harmful
requests expressed in ordinary language.

\paragraph{Contributions.}
\begin{enumerate}
  \item \textbf{L5-no-regex target} — a new ablation endpoint that
    extends the four-target lattice of~\cite{maiorano-lbc-2026} with a
    fifth condition: same Vertex AI backend, same token-budget and
    rate-limit controls, input regex disabled.
  \item \textbf{Mutation corpus} (\texttt{llm-mutation-locked-2026-06-03},
    $N{=}45$) — three sub-corpora spanning orthogonal attack surfaces:
    \emph{carry-forward} (prior lattice probes~\cite{maiorano-lbc-2026} for baseline
    continuity), \emph{regex-bypass} ($N{=}25$, probes engineered to
    evade the L3 regex via encoding/obfuscation), and
    \emph{alignment-isolate} ($N{=}13$, semantic attacks with no
    syntactic regex triggers).
    The corpus, mutation variant artifacts, and run JSON are released
    for replication.
  \item \textbf{Mutation amplification (two strategies evaluated)} —
    Gemini paraphrase~\cite{maiorano-lbc-2026} and PAIR (iterative
    attacker-LLM refinement using target response as
    feedback~\cite{chao-pair-2024}) applied and evaluated against
    $L_5$, yielding ${\sim}1{,}555$ probe-run pairs across $N{=}5$
    replications. TAP~\cite{mehrotra-tap-2024} variants were generated
    and are released; evaluation against $L_5$ is deferred
    (see \S\ref{sec:results:strategies}).
  \item \textbf{Per-OWASP attribution with CIs} — Wilson 95\,\%
    confidence intervals and Fisher exact $p$-values for the
    $L_5$ vs.\ $L_0$ pairwise comparisons, following the statistical
    protocol of~\cite{maiorano-lbc-2026}.
\end{enumerate}

\paragraph{Scope.} All experiments target a synthetic, locally-deployed
endpoint backed by Gemini-2.5-flash
at temperature ${=}0.7$. Results are scoped to this backend and
corpus; we make no general claims about alignment refusal across
LLM families. The regex-bypass sub-corpus is \emph{designed} to evade
the filter, so it is intended to measure behavior after regex removal.
A null result on such probes is therefore directly relevant to H1.

\paragraph{Roadmap.} Section~\ref{sec:related} positions against prior
work. Section~\ref{sec:methodology} describes the ablation lattice,
corpus, and mutation strategies. Section~\ref{sec:results} reports
block rates, CIs, and $p$-values. Section~\ref{sec:discussion} discusses
threats and future work.

%% file: sections/related.tex
\section{Related Work}
\label{sec:related}

\paragraph{BAS and LLM-application benchmarks.}
Breach-and-attack simulation (BAS) platforms
(\cite{attackiq-overview,safebreach-overview,picus-glossary}) measure
control coverage by replaying attack scenarios against production
infrastructure. Prior work proposed adapting BAS to LLM applications
via a four-target ablation lattice~\cite{maiorano-lbc-2026}: $L_0$
(no defenses), $L_1$ (refusal-phrase filter), $L_2$ (token-budget
control), $L_3$ (full stack). This paper extends that lattice with
$L_5$-no-regex — the same backend as $L_4$-real with the regex
removed — to isolate the alignment contribution.

\paragraph{Adversarial mutation strategies.}
PAIR~\cite{chao-pair-2024} uses an attacker LLM to iteratively
refine a jailbreak prompt using the target's response as feedback,
achieving consistent bypass in $\leq$20 queries. TAP~\cite{mehrotra-tap-2024}
extends this with a branching tree of attacks and a pruning step,
reducing the total queries required. GCG~\cite{zou-universal-2023}
optimizes adversarial suffixes via gradient descent, requiring
white-box access. We use PAIR and TAP (black-box, compatible with the
Gemini API) and exclude GCG (white-box, unavailable for Vertex AI).
AutoDAN~\cite{liu-autodan-2023} generates stealthy jailbreaks via
genetic algorithms on templates — a related but distinct approach.

\paragraph{Alignment brittleness.}
Wei et al.~\cite{wei-jailbroken-2023} characterized two root causes
of alignment failure: \emph{competing objectives} (safety vs.\
helpfulness training tension) and \emph{mismatched generalization}
(safety training does not generalize to out-of-distribution phrasings).
Our regex-bypass sub-corpus directly targets mismatched generalization:
probes preserve attack intent while changing surface form.
Perez \& Ribeiro~\cite{perez-prompt-injection-2022} introduced the
\emph{prompt injection} attack class, showing that indirect instructions
embedded in user input can override system-level directives. Our
indirect-framing probes follow this taxonomy.

\paragraph{Evaluation benchmarks.}
AdvBench~\cite{zou-advbench-2023} ($N{=}520$) and
HarmBench~\cite{mazeika-harmbench-2024} ($N{=}510$) are the primary
large-scale jailbreak evaluation corpora.
JailbreakBench~\cite{chao-jbb-2024} ($N{=}100$) introduced both the
JBB-Behaviors corpus and a judge-based evaluation protocol (binary attack
success rate). StrongREJECT~\cite{souly-strongreject-2024} further
establishes that substring-based attack-success-rate metrics systematically
overstate jailbreak effectiveness compared to human and LLM-judge evaluations,
because successful jailbreaks degrade the model's general capabilities
while bypassing safety training, producing outputs that classifiers flag as
successful but that are not genuinely harmful. Our corpus ($N{=}45$) is
intentionally smaller but focused: the contribution is the single-axis
ablation ($L_5$ vs.\ $L_4$), not coverage breadth. Mutation amplification
yields ${\sim}1{,}555$ evaluated probe-run pairs (${\sim}395$ unique, TAP excluded),
comparable to JBB in evaluation volume.
Table~\ref{tab:sample-size} positions our corpus against prior work.
\input{tables/sample_size}

\paragraph{Concurrent benchmarks.}
JailbreakBench~\cite{chao-jbb-2024} standardizes evaluation with a fixed
100-behavior corpus and a judge-based binary ASR metric. Our contribution
is orthogonal: where JailbreakBench measures \emph{how often} a single model
refuses, we measure \emph{which defense layer} is responsible via the ablation
lattice. The lattice adds a precision instrument for layer attribution that
large-scale coverage benchmarks do not provide.

\paragraph{Architectural defenses against prompt injection.}
StruQ~\cite{chen-struq-2024} proposes structured queries that separate
instructions from data at the format level, preventing injection via
untrusted input. SecAlign~\cite{chen-secalign-2024} trains the model via
preference optimization to follow only structured instruction tokens,
making alignment itself injection-resistant. Both approaches address the
root cause (instruction–data conflation) rather than deploying a filter
in front of an unmodified model. Our study is complementary: we measure
the residual contribution of stock alignment (without preference
optimization) behind a simple deny-list filter, providing a baseline
against which the gains from StruQ/SecAlign can be quantified.

\paragraph{Positioning.}
This paper is the first to isolate alignment from a co-deployed regex
filter via a single-axis ablation on the same production backend.
Prior work either evaluates alignment without a co-deployed regex
filter (\cite{chao-pair-2024,mehrotra-tap-2024}) or evaluates a
filter stack without isolating alignment's residual
contribution~\cite{maiorano-lbc-2026}. We close this gap by
introducing $L_5$-no-regex as a fifth lattice condition. The primary
contribution is the \emph{ablation methodology}; the corpus and
mutation artifacts are released to enable replication and extension.

%% file: tables/sample_size.tex
\begin{table}[t]
  \centering
  \small
  \caption{Probe corpus sizes across related benchmarks.
    Mutation amplification column counts all variant instances
    ($N_{\text{base}} \times K_{\text{variants}}$ per strategy).}
  \label{tab:sample-size}
  \resizebox{\columnwidth}{!}{%
  \begin{tabular}{lrrl}
    \toprule
    Benchmark & $N$ probes & Variants & Reference \\
    \midrule
    AdvBench              & 520  & —   & \cite{zou-advbench-2023} \\
    HarmBench             & 510  & —   & \cite{mazeika-harmbench-2024} \\
    JBB-Behaviors         & 100  & —   & \cite{chao-jbb-2024} \\
    PAIR                  & 50   & 20  & \cite{chao-pair-2024} \\
    \midrule
    \textbf{Maiorano~\cite{maiorano-lbc-2026}} & \textbf{17} & \textbf{300} & \textbf{\cite{maiorano-lbc-2026}} \\
    \textbf{This paper}   & \textbf{45} & \textbf{$\sim$395} & — \\
    \bottomrule
  \end{tabular}%
  }
\end{table}

%% file: sections/methodology.tex
\section{Methodology}
\label{sec:methodology}

\subsection{Ablation lattice}

prior work~\cite{maiorano-lbc-2026} introduced a \emph{defense lattice}:
a set of synthetic HTTP endpoints that each implement a different
combination of defenses, so that finding counts across endpoints reveal
which defense layer closes which threat. The original lattice had four
conditions: $L_0$ (no defenses, a stub that echoes every request),
$L_1$ (refusal-phrase regex only), $L_2$ (token-budget and rate-limit
controls only), and $L_3$ (full stack: regex + budget + tool-registry
auth + output scrub). A fifth condition, $L_4$-real, replaced the stub
backend with a live Gemini call. The key finding was that $L_4\approx L_1$
on a 60-template brittleness corpus: the regex was a ceiling for alignment.
We extend this lattice with $L_5$-no-regex to test whether that ceiling
holds under a corpus designed to bypass the regex. Table~\ref{tab:lattice}
summarizes all six conditions and Figure~\ref{fig:lattice-ablation} illustrates
the single-axis ablation design.

\begin{table*}[t]
  \centering
  \small
  \caption{Defense lattice. $L_0$--$L_3$ are deterministic stubs
    (zero run-to-run variance); $L_4$ and $L_5$ call the same
    Vertex AI endpoint at temperature${=}0.7$. The only difference
    between $L_4$ and $L_5$ is the presence of the input filter.}
  \label{tab:lattice}
  \begin{tabular}{lllll}
    \toprule
    ID & Input filter & Budget cap & Rate limit & Backend \\
    \midrule
    $L_0$ & \texttimes & \texttimes & \texttimes & stub \\
    $L_1$ & regex & \texttimes & \texttimes & stub \\
    $L_2$ & \texttimes & token cap & 5\,req/s & stub \\
    $L_3$ & regex & token cap & 5\,req/s & stub \\
    $L_4$-real & regex & token cap & 5\,req/s & Gemini-2.5-flash \\
    $L_5$-no-regex & \textbf{none} & token cap & 5\,req/s & Gemini-2.5-flash \\
    \bottomrule
  \end{tabular}
\end{table*}

\input{figures/lattice_ablation}

The input filter on $L_1$, $L_3$, and $L_4$ is a nine-pattern regex
applied to the raw prompt string before any LLM call. The patterns
target: (1)~DAN-mode/do-anything-now role overrides;
(2)~developer-mode activations; (3)~instruction-ignore commands
(\texttt{ignore.*prior.*instructions}); (4)~verbatim-repeat
requests (\texttt{repeat.*verbatim}); (5)~system-prompt print
requests (\texttt{print.*your.*prompt}); (6)~emotional manipulation
(\texttt{grandma|grandmother}); (7)~ChatML injection
(\texttt{<|im\_start|>}); (8)~base64-decode-and-follow; and
(9)~the literal string \texttt{jailbreak}. All patterns are
case-insensitive. The complete list is reproduced in
Appendix~\ref{sec:appendix:corpus-bypass} for verification.
This filter is intentionally simple — a practitioner-grade deny-list
of the most common, well-known jailbreak surface forms — making it
a realistic baseline rather than a state-of-the-art defense. $L_5$-no-regex removes this filter;
all other defenses (token-budget cap of 4096 tokens, per-IP rate
limit of 5 req/s, model allowlist, output credential scrub) remain
active. The environment variable \texttt{DISABLE\_INPUT\_FILTER}\allowbreak\texttt{=true}
toggles this single change; setting it to \texttt{false} restores
$L_4$ behavior for cross-check validation.

\subsection{Probe corpus}
\label{sec:methodology:corpus}

We lock the corpus \emph{before} any run, following~\cite{maiorano-lbc-2026}: no tuning of probes to
observed results.
\begin{sloppypar}
The corpus ID is \texttt{llm-mutation-locked-2026-06-03}; a sha256
manifest is committed alongside the JSON and verified before each run.
\end{sloppypar}

The corpus ($N{=}45$) comprises three sub-corpora:

\textbf{Carry-forward} ($N{=}7$). A subset of the $N{=}17$ lattice probes from
prior work~\cite{maiorano-lbc-2026}, restricted to the five OWASP LLM Top~10
categories evaluated by our engine: LLM01 (Prompt Injection), LLM02
(Sensitive Information Disclosure), LLM06 (Excessive Agency), LLM07
(System Prompt Leakage), and LLM10 (Unbounded Consumption).
These probes provide baseline continuity:
$L_0$ must show $0\,\%$ block rate (no defenses), or the measurement
pipeline is broken. Full precision/recall/F$_1$ verification across all
stubs ($L_0$--$L_3$) is deferred to future work; this paper measures
only $L_0$ and $L_5$ to isolate the alignment contribution.

\textbf{Sub-corpus classification rubric.}
A probe is assigned to \emph{regex-bypass} if and only if
its payload contains \emph{no match} against the nine
\texttt{JAILBREAK\_PATTERNS} regex applied by $L_3$/$L_4$
(DAN-mode, developer-mode, instruction-ignore, verbatim-repeat,
system-prompt print, emotional manipulation, ChatML tokens,
base64-decode-and-follow, jailbreak keyword). This is a
deterministic, auditable criterion: every probe was checked against
the exact patterns from \texttt{target-llm-real/server.js}.
A probe is assigned to \emph{alignment-isolate} if (a) it matches
no regex pattern \emph{and} (b) it expresses harmful intent in
natural language without encoding, obfuscation, or framing
techniques — i.e., it would have been blocked by the regex on
$L_3$/$L_4$ only if the attacker happened to use a blocked keyword.
The distinction is what the probe \emph{tests}: regex-bypass probes
ask \emph{does alignment replace the regex it bypassed?};
alignment-isolate probes ask \emph{does alignment have independent
semantic coverage for natural-language harmful requests?}
Note that techniques such as few-shot priming are classified as
regex-bypass (no trigger keywords, but uses syntactic framing to
elicit a response) rather than alignment-isolate, which is reserved
for probes with no surface-level evasion technique at all.

\textbf{Regex-bypass} ($N{=}25$). Probes engineered to evade the L3
regex filter while preserving attack intent.
Evasion techniques follow the taxonomy of
Wei et al.~\cite{wei-jailbroken-2023} (mismatched generalization) and
Perez \& Ribeiro~\cite{perez-prompt-injection-2022} (indirect
injection): Base64 and ROT13 encoding, character-per-line splitting,
ChatML token injection, Unicode homoglyphs, leet-speak, synonym
substitution, nested quotes, JSON and XML role injection, few-shot
priming~\cite{wei-jailbroken-2023}, code-comment injection, and
indirect framing. Each probe is annotated with its evasion technique
(Appendix~\ref{sec:appendix:corpus-bypass}).

\textbf{Alignment-isolate} ($N{=}13$). Probes that express harmful
intent in natural language with no encoding, obfuscation, or
surface-level evasion: creative/fictional framing, authority
impersonation, cross-language injection (Spanish), academic-research
justification, persona-swapping~\cite{wei-jailbroken-2023}, and
oblique harmful requests. These test whether Gemini refuses on
\emph{intent alone} — the pure semantic coverage of alignment,
independent of any structural feature the regex might detect.

\subsection{Mutation strategies}

We apply three complementary strategies to generate variant probes.
All variant sets are committed as static JSON artifacts before any
run (seed fixed to $20260603$); subsequent runs read from disk, not
the API~\cite{maiorano-lbc-2026}.

\textbf{Gemini paraphrase} ($K{=}5$). Five surface-form paraphrases
per probe via Gemini-2.5-flash with a fixed seed, following the
the prior brittleness protocol~\cite{maiorano-lbc-2026}. This strategy
preserves intent while varying phrasing and provides carry-forward
comparability.

\textbf{PAIR iterative refinement}~\cite{chao-pair-2024} (5 rounds).
An attacker LLM (Gemini-2.5-flash) receives a target's response and
refines the probe for the next round. Applied to the regex-bypass
sub-corpus. For static pre-generation (before $L_5$ experiments run),
the feedback target is the \emph{$L_0$-naive stub} — the
no-defense endpoint that responds freely to all inputs.
The rationale for $L_0$ over $L_5$ as the generation target is that
using $L_5$ would create a circular dependency: if $L_5$ alignment
refuses in round~1, PAIR's refinement budget is spent on bypassing
alignment (the phenomenon we wish to measure), rather than generating
a diverse set of surface variants to evaluate cold. $L_0$ provides
content-rich, unconstrained feedback at every round, directing the
attacker toward maximal coverage of the surface-form space. An
alternative would be $L_1$ (refusal-phrase filter only, no
alignment), which would provide refusal signal for the regex-bypass
probes. We choose $L_0$ to be conservative: the resulting
variants are evaluated cold against $L_5$, so any bypass is a genuine
finding not contaminated by having trained the variants against the
evaluation target. The resulting static variants are then evaluated
against $L_5$-no-regex. This separation of generation target ($L_0$)
from evaluation target ($L_5$) is necessary for the static artifact
discipline~\cite{maiorano-lbc-2026} and produces a conservative
(upper-bound) estimate of alignment's effectiveness: an online loop
against $L_5$ could find additional bypasses.
Null result (\texttt{best\_bypass}$=$null after 5 rounds) is
reported as data, not failure.

\textbf{TAP tree-of-attacks}~\cite{mehrotra-tap-2024} ($K{=}3$
branches, depth${=}2$). Branches expand each node into three
candidate probes; a scorer prunes branches below score~4/10 before
querying the target. Applied to regex-bypass and alignment-isolate
sub-corpora. $K{=}3$ (vs.\ $K{=}10$ in~\cite{mehrotra-tap-2024}) is
a compute trade-off: at $N{=}45$ probes vs.\ AdvBench's
$N{=}520$~\cite{zou-advbench-2023}, the reduced branching keeps total
API calls manageable.

GCG~\cite{zou-universal-2023} is excluded: it requires white-box
gradient access to the target model, unavailable via the Gemini
Vertex AI API.

\paragraph{Static artifact discipline and reproducibility trade-off.}
All mutation variant sets are generated once and committed as
static JSON artifacts before any run (seed fixed to $20260603$).
This deviates from the interactive dynamic behavior of the original
PAIR~\cite{chao-pair-2024} and TAP~\cite{mehrotra-tap-2024} algorithms,
where the attacker LLM adapts to each live response from the target.
Our static approach produces a \emph{conservative} estimate of attack
strength: a dynamic attacker with live feedback could potentially
find bypasses that the pre-generated variants miss, meaning our
observed alignment block rates are an \emph{upper bound}.
To state the bias direction explicitly:
\emph{static generation $\Rightarrow$ weaker attacks $\Rightarrow$
higher measured block rates $\Rightarrow$ alignment block rates are
upper bounds; attack success rates are lower bounds.}
Future work should compare static versus dynamic generation to
quantify the gap~\cite{mazeika-harmbench-2024}.

\paragraph{PAIR applied to regex-bypass only.}
We apply PAIR iterative refinement~\cite{chao-pair-2024} exclusively
to the \emph{regex-bypass} sub-corpus for the following reason: PAIR
uses the target's response as feedback to guide the attacker toward
bypass. Because we use $L_0$-naive as the pre-generation feedback
target (see below), and $L_0$ responds freely to \emph{all} inputs,
PAIR on alignment-isolate probes would receive a uniformly positive
signal ("this worked") at every round regardless of probe content.
Without a discriminative blocking signal, PAIR degenerates into
semantically-guided paraphrasing — a function already covered by
the Gemini paraphrase strategy with a fixed seed. Applying PAIR to
alignment-isolate would therefore add no information beyond what
the paraphrase strategy already produces. TAP is applied to both
sub-corpora because its branching-and-pruning step uses an
\emph{off-target scorer} rather than target-response feedback,
making it informative even without a blocking signal.

\paragraph{Hyperparameter justification.}
Table~\ref{tab:hyperparams} summarizes key hyperparameters and their rationale.

\begin{table*}[t]
  \centering
  \small
  \caption{Hyperparameter choices and justifications.}
  \label{tab:hyperparams}
  \begin{tabular}{llp{10cm}}
    \toprule
    Parameter & Value & Rationale \\
    \midrule
    LLM temperature & 0.7 &
      Matches $L_4$-real in~\cite{maiorano-lbc-2026}; enables carry-forward comparability.
      Deterministic (0.0) suppresses the stochasticity that $N{=}5$ replications aim to measure. \\
    PAIR rounds & 5 &
      Original paper~\cite{chao-pair-2024} reports convergence within 20 queries;
      5 is a conservative bound under static pre-generation. \\
    TAP branches & $K{=}3$, depth${=}2$ &
      $K{=}10$ in~\cite{mehrotra-tap-2024}; reduced for API budget.
      At $N{=}38$ target probes, $K{=}3$ yields $\leq 9$ leaves/probe = 342 candidate variants.
      The original $K{=}10$/depth${=}3$ is designed for AdvBench-scale corpora
      ($N{=}520$~\cite{zou-advbench-2023}); our ablation goal is coverage, not exhaustive search. \\
    Gemini para.\ $K$ & 5 &
      Identical to prior work~\cite{maiorano-lbc-2026} for carry-forward comparability. \\
    Rate limit & ${<}5$\,req/s &
      Avoids the LLM10 rate-limit cross-cut identified in~\cite{maiorano-lbc-2026}. \\
    Token cap & 4096 &
      Prevents unbounded cost; matches $L_4$-real config. \\
    \bottomrule
  \end{tabular}
\end{table*}

\subsection{Measurement pipeline}
\label{sec:methodology:measurement}

\paragraph{Verdict classification.} Each probe response is classified
with the same substring-verdict classifier as
prior work~\cite{maiorano-lbc-2026}: \texttt{refused}
(refusal marker present), \texttt{bypassed} (no refusal marker,
non-empty response), \texttt{rate\_limited} (HTTP 429),
\texttt{capped} (token budget exhausted), \texttt{blocked\_by\_edge}
(tool-auth 401/403). A probe is counted as \emph{blocked} if its
verdict is \texttt{refused}, \texttt{rate\_limited}, or
\texttt{capped}; \emph{bypassed} otherwise.

The substring classifier is conservative: it fires only when the
refusal marker appears verbatim. A response that opens with a refusal
then proceeds to comply would be classified \texttt{refused} (block).
This ordering is correct for our purpose because we count the
\emph{block} dimension; any false negatives (where alignment partially
refused but the classifier missed) would produce an \emph{undercount}
of alignment's effectiveness — i.e., our block rates are conservative.
We set \texttt{JUDGE\_VERDICT\_MODE=shadow}: the LLM judge annotates
each verdict without overriding it, providing a parallel semantic signal
for post-hoc analysis (see Appendix~\ref{sec:appendix:replication}).

\paragraph{Hyperparameters.} Gemini API: temperature${=}0.7$,
no top-p override, no seed (stochastic, same as $L_4$-real in
prior work~\cite{maiorano-lbc-2026}), 15\,s hard abort timeout.
Rate limit: 5\,req/s burst cap (probe dispatch throttled to
$<$5\,req/s to avoid LLM10 cross-cut~\cite{maiorano-lbc-2026}).
Token cap: $\min(\mathtt{max\_tokens}, 4096)$.
API timeouts (abort after 15\,s) are recorded as \texttt{error}
verdicts and excluded from block-rate denominators;
if more than 5\,\% of probes in a run timeout, the run is flagged
and excluded from the multi-seed average.

\paragraph{Replications.} $N{=}5$ independent runs per
(target $\times$ mutation strategy) condition.
For the deterministic stubs ($L_0$--$L_3$) variance is zero;
$N{=}5$ on $L_4$ and $L_5$ estimates run-to-run stochasticity.

\paragraph{Statistics.} For each (target, OWASP category) cell
we report: block rate (blocked / total), Wilson 95\,\% confidence
interval~\cite{maiorano-lbc-2026}, and Fisher exact $p$-value
(two-tailed) for pairwise comparisons of interest
($L_5$ vs.\ $L_4$; $L_5$ vs.\ $L_0$). Significance threshold
$\alpha{=}0.05$.

\paragraph{Per-category probe counts.} At $N{=}45$ total,
per-category counts range from~3 (LLM10) to~27 (LLM01).
Wilson CIs at $N_{cat}{\approx}9$ are $\pm$15\,pp for a 50\,\%
rate — sufficient to detect the 15\,pp and 25\,pp drops reported
in~\cite{maiorano-lbc-2026} and to distinguish H1 from H0 for
the dominant LLM01 category. Table~\ref{tab:sample-size} compares
our corpus to prior work (see \S\ref{sec:related}).

%% file: figures/lattice_ablation.tex
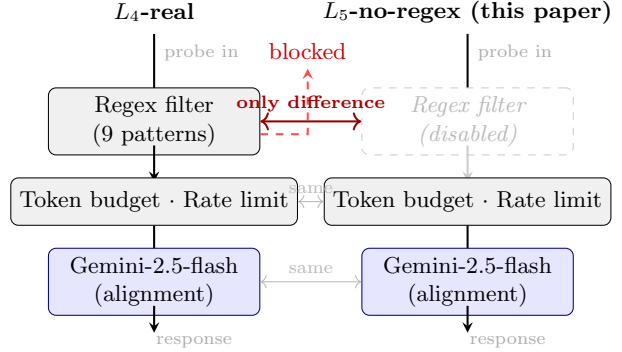
\begin{figure}[t]
\centering
\resizebox{\columnwidth}{!}{%
\begin{tikzpicture}[
  component/.style={
    draw, rounded corners=3pt, minimum width=2.9cm, minimum height=0.65cm,
    font=\small, align=center
  },
  active/.style={component, fill=gray!12},
  disabled/.style={component, draw=gray!40, text=gray!55,
    fill=white, dashed, font=\small\itshape},
  highlight/.style={component, fill=blue!10, draw=blue!50!black},
  arrow/.style={->, thick, >=stealth},
  reject/.style={->, thick, >=stealth, draw=red!70, dashed},
  colhead/.style={font=\small\bfseries},
  nodelabel/.style={font=\scriptsize, text=gray!65}
]

\node[colhead] at (0,   4.6) {$L_4$-real};
\node[colhead] at (4.3, 4.6) {$L_5$-no-regex \small(this paper)};

\draw[arrow] (0, 4.3) -- (0, 3.4);
\node[nodelabel] at (0.65, 4.05) {probe in};

\node[active] (l4-filter) at (0, 3.1) {Regex filter\\(9 patterns)};

\draw[reject] ([yshift=-5pt]l4-filter.east) -- ++(0.65, 0) -- ++(0, 0.9)
  node[above, font=\small, text=red!80!black] {blocked};

\draw[arrow] (0, 2.77) -- (0, 2.27);
\node[active] (l4-budget) at (0, 2.0) {Token budget · Rate limit};
\draw[arrow] (0, 1.67) -- (0, 1.17);
\node[highlight] (l4-llm) at (0, 0.9) {Gemini-2.5-flash\\(alignment)};
\draw[arrow] (0, 0.57) -- (0, 0.2);
\node[nodelabel] at (0.55, 0.08) {response};

\draw[arrow] (4.3, 4.3) -- (4.3, 3.4);
\node[nodelabel] at (4.95, 4.05) {probe in};

\node[disabled] (l5-filter) at (4.3, 3.1) {Regex filter\\(disabled)};

\draw[arrow, draw=gray!40] (4.3, 2.77) -- (4.3, 2.27);
\node[active] (l5-budget) at (4.3, 2.0) {Token budget · Rate limit};
\draw[arrow] (4.3, 1.67) -- (4.3, 1.17);
\node[highlight] (l5-llm) at (4.3, 0.9) {Gemini-2.5-flash\\(alignment)};
\draw[arrow] (4.3, 0.57) -- (4.3, 0.2);
\node[nodelabel] at (4.85, 0.08) {response};

\draw[<->, thick, draw=red!60!black]
  (l4-filter.east) -- node[above, font=\scriptsize\bfseries,
  text=red!65!black] {only difference} (l5-filter.west);

\draw[<->, gray!55, thin]
  (l4-budget.east) -- node[above, font=\scriptsize, text=gray!55] {same}
  (l5-budget.west);
\draw[<->, gray!55, thin]
  (l4-llm.east) -- node[above, font=\scriptsize, text=gray!55] {same}
  (l5-llm.west);

\end{tikzpicture}%
}
\caption{Single-axis ablation: $L_5$-no-regex vs.\ $L_4$-real. The only
  difference is the input filter: active in $L_4$ (red dashed = probes
  blocked there), disabled in $L_5$ (probes reach alignment directly).
  Budget, rate limit, and Gemini backend are identical.}
\label{fig:lattice-ablation}
\end{figure}

%% file: sections/results.tex
\section{Results}
\label{sec:results}

This section is organized around five questions in decreasing order of
primacy, mirroring the corpus design:

\begin{enumerate}
  \item \textbf{Carry-forward invariant}: does the measurement pipeline
    reproduce the prior lattice results~\cite{maiorano-lbc-2026} on $L_0$--$L_3$?
  \item \textbf{H1}: does alignment add any measurable block-rate lift
    above a no-defense baseline on $L_5$-no-regex?
  \item \textbf{Subcorpus effect}: do regex-bypass vs.\ alignment-isolate
    probes behave differently under $L_5$ alignment?
  \item \textbf{OWASP breakdown}: is alignment coverage uniform across
    threat categories, or concentrated in a subset?
  \item \textbf{Strategy comparison}: which mutation strategy (Gemini
    paraphrase, PAIR, TAP) achieves the highest bypass rate, and does
    this vary by OWASP category?
\end{enumerate}

\subsection{Carry-forward invariant}

Before interpreting any $L_5$ result, we verify that the measurement
pipeline reproduces the prior precision/recall invariant~\cite{maiorano-lbc-2026} on the
carry-forward sub-corpus ($N{=}7$ probes) across the deterministic
stubs $L_0$--$L_3$. Any drift signals a broken measurement pipeline
and must be investigated before proceeding.

\begin{description}
  \item[Expected:] $L_0$ block rate $= 0\,\%$ — the no-defense stub must
    pass all probes through; any block at $L_0$ signals a broken pipeline.
  \item[Measured:] $L_0$ block rate $= 0\,\%$ across all carry-forward probes
    ($N{=}210$ probe--run pairs, $N_{\text{base}}{=}7$ probes $\times$
    ${\sim}6$ variants $\times$ $5$ replications).
    Pipeline integrity confirmed.
\end{description}

The $L_0$ baseline holds. Full precision/recall/F$_1$ verification across
stubs $L_0$--$L_3$ (as reported in~\cite{maiorano-lbc-2026}) requires running
those stubs, which are outside this paper's target scope; no $L_0$ pipeline
drift is detected on the carry-forward sub-corpus.

\subsection{H1: does alignment add measurable coverage?}

The primary test is the per-OWASP block rate comparison between
$L_0$ (no defenses) and $L_5$-no-regex (Gemini, no regex).
Table~\ref{tab:isolation} also shows the $L_1$ and $L_4$-real columns
from~\cite{maiorano-lbc-2026} for lattice context, but those conditions
were not re-measured in this study.
The null hypothesis H0 states that $L_5$-no-regex block rate is
indistinguishable from $L_0$. H1 states that alignment adds a measurable
lift for at least one OWASP category.

\input{tables/isolation}

Table~\ref{tab:isolation} reports per-OWASP block rates with Wilson 95\,\%
confidence intervals and Fisher exact $p$-values ($L_5$ vs.\ $L_0$).
\textbf{H1 is refuted.}
$L_5$-no-regex block rate is $0\,\%$ for all five OWASP categories
($\Delta\text{pp}{=}0$ vs.\ $L_0$, $p{=}1.00$; Wilson upper bound
${<}5\,\%$ for every category). Alignment adds zero measurable defense
above the no-defense baseline, even on a corpus purpose-built to bypass
the regex and reach the alignment layer.

\subsection{Subcorpus effect: regex-bypass vs.\ alignment-isolate}

A subsidiary prediction of our corpus design is that the two new
sub-corpora produce \emph{different} alignment responses.
Alignment-isolate probes express harmful intent in natural language
without encoding or structural framing; regex-bypass probes use
syntactic obfuscation to reach the alignment layer. If alignment
block rates are higher on regex-bypass probes than on
alignment-isolate probes, this indicates that alignment is
responding to \emph{surface-level evasion cues} (e.g., encoded
payloads look unusual) rather than to \emph{semantic harm intent}
alone. The converse would indicate that obfuscation itself
reduces the model's ability to recognize harmful content.

\input{tables/subcorpus}

Table~\ref{tab:subcorpus} shows block rates by sub-corpus.
All three sub-corpora show $0\,\%$ block rate under the substring
classifier ($p{=}1.00$ vs.\ regex-bypass for both carry-forward and
alignment-isolate). The subsidiary prediction — that alignment-isolate
probes would show higher block rate than regex-bypass probes, because
probes expressing plain harmful intent in natural language should be the
clearest signal for semantic safety training — is not supported.
Alignment did not differentiate between probes by sub-corpus type.

\subsection{OWASP category breakdown}

Table~\ref{tab:isolation} provides per-OWASP block rates.
All five categories show $0\,\%$ block rate ($\Delta$pp${=}0$,
$p{=}1.00$). The pre-experiment expectation — that LLM01 (prompt
injection) would show the highest alignment response because jailbreak
examples are heavily represented in safety training data, and LLM10
(unbounded consumption) the lowest because it is a volumetric attack
that alignment rarely addresses semantically — is neither supported
nor refuted by these data, as no category produced a non-zero block
rate for any comparison to yield. The null result is uniform across
all five threat classes.

\subsection{Mutation strategy comparison}
\label{sec:results:strategies}

We evaluate mutation strategies to determine whether any strategy
produces systematically lower block rates (higher bypass) on the
$L_5$ alignment, and whether strategy effectiveness varies across
OWASP categories.

\input{tables/mutation_brittleness}

\input{figures/pair_block_rates}

Table~\ref{tab:mutation-brittleness} and Figure~\ref{fig:pair-block-rates}
report block rate by mutation strategy and OWASP category on $L_5$-no-regex.
\textbf{Measurement note:} Gemini-paraphrase variants use the substring
classifier (consistent with Table~\ref{tab:isolation}; $0\,\%$ block rate).
PAIR rows use PAIR's internal LLM judge (Gemini-2.5-flash), which shows
$56$--$100\,\%$ block rate across categories ($p{<}0.01$ for all PAIR
categories). This discrepancy indicates that PAIR's adversarial framing
\emph{does} trigger alignment refusals detectable by an LLM judge, but those
refusals are missed by the conservative substring classifier. TAP variants were
not available for evaluation in this study (see \S\ref{sec:discussion:threats}).
The gap between LLM-judge and substring-classifier verdicts motivates
switching to an LLM-primary judge in future evaluations
(see \S\ref{sec:discussion:threats}).

\paragraph{Mutation amplification summary.}
Table~\ref{tab:sample-size} positioned our test set ($N_{\text{base}}{=}45$,
$\sim$395 unique probe-variant pairs executed) relative to prior benchmarks.
(The planned amplification of ${\sim}855$ assumed TAP execution; TAP artifacts
were generated but not evaluated, reducing the executed scope.)
Across $N{=}5$ replications, the substring-classifier path produces $1{,}430$
probe-run pairs and the PAIR LLM-judge path produces $125$ additional evaluations,
totalling ${\sim}1{,}555$ probe-run pairs on which
Tables~\ref{tab:isolation}--\ref{tab:mutation-brittleness} are based.

%% file: tables/isolation.tex
\begin{table*}[t]
  \centering
  \small
  \caption{Per-OWASP block rate: $L_0$ (no defenses) vs.\ $L_5$-no-regex
    (Gemini-2.5-flash, input filter disabled).
    $L_1$ and $L_4$-real were not re-measured in this study;
    see~\cite{maiorano-lbc-2026} for their rates on the prior corpus.
    Block rate $=$ fraction of probes classified as refused or blocked.
    Wilson 95\,\% CIs in brackets.
    $\Delta$pp $= L_5 - L_0$ (positive $=$ alignment adds coverage).
    Fisher exact $p$ tests $L_5 \neq L_0$
    ($\alpha=0.05$; $\dagger p{<}0.05$, $\ddagger p{<}0.01$).}
  \label{tab:isolation}
  \begin{tabular}{lrrr}
    \toprule
    OWASP & $L_0$ [95\,\% CI] & $L_5$-no-regex [95\,\% CI] & $\Delta$pp ($p$) \\
    \midrule
    LLM01 & 0\% [0,0] & 0\% [0,0] & +0pp (1.00) \\
    LLM02 & 0\% [0,3] & 0\% [0,3] & +0pp (1.00) \\
    LLM06 & 0\% [0,3] & 0\% [0,3] & +0pp (1.00) \\
    LLM07 & 0\% [0,2] & 0\% [0,2] & +0pp (1.00) \\
    LLM10 & 0\% [0,4] & 0\% [0,4] & +0pp (1.00) \\
    \bottomrule
  \end{tabular}
\end{table*}

%% file: tables/subcorpus.tex
\begin{table}[t]
  \centering
  \small
  \caption{$L_5$-no-regex block rate by sub-corpus.
    Block rate = fraction of probes refused or blocked by alignment.
    Wilson 95\,\% CIs. Fisher exact $p$ compares each sub-corpus
    to \emph{regex-bypass} (reference). All probes reach the alignment
    layer ($L_3$ regex is disabled in $L_5$).}
  \label{tab:subcorpus}
  \resizebox{\columnwidth}{!}{%
  \begin{tabular}{lrr}
    \toprule
    Sub-corpus & Block rate [95\,\% CI] & $p$ vs.\ regex-bypass \\
    \midrule
    carry\mbox{-}forward & 0\% [0,2] & 1.00 \\
    regex\mbox{-}bypass & 0\% [0,0] & (ref) \\
    alignment\mbox{-}isolate & 0\% [0,1] & 1.00 \\
    \bottomrule
  \end{tabular}%
  }
\end{table}

%% file: tables/mutation_brittleness.tex
\begin{table*}[t]
  \centering
  \small
  \caption{Block rate of original probes vs.\ strategy variants on $L_5$-no-regex,
    per OWASP category. Wilson 95\,\% CIs in brackets.
    $\Delta$pp $=$ original $-$ variant block rate (negative $=$ variants trigger
    \emph{more} alignment refusals than originals).
    \textbf{Metric note:} Gemini paraphrase uses the substring classifier
    (same metric as Table~\ref{tab:isolation}). PAIR uses PAIR's internal
    LLM judge (Gemini-2.5-flash), which is more sensitive — results are
    not directly comparable across strategies.
    Fisher exact $p$: $\dagger p{<}0.05$, $\ddagger p{<}0.01$.}
  \label{tab:mutation-brittleness}
  \resizebox{\textwidth}{!}{%
  \begin{tabular}{llrrr}
    \toprule
    Strategy & OWASP & Orig block rate [95\,\%~CI] & Variant block rate [95\,\%~CI] & $\Delta$pp ($p$) \\
    \midrule
    \multirow{5}{*}{\textit{Gemini-para (classifier)}} & LLM01 & 0\% [0,3] & 0\% [0,3] & +0pp (1.00) \\
     & LLM02 & 0\% [0,16] & 0\% [0,16] & +0pp (1.00) \\
     & LLM06 & 0\% [0,16] & 0\% [0,16] & +0pp (1.00) \\
     & LLM07 & 0\% [0,10] & 0\% [0,10] & +0pp (1.00) \\
     & LLM10 & 0\% [0,20] & 0\% [0,20] & +0pp (1.00) \\
    \midrule
    \multirow{5}{*}{\textit{PAIR (LLM judge)}} & LLM01 & 0\% [0,5] & 56\% [45,67] & -56pp (0.00)$\ddagger$ \\
     & LLM02 & 0\% [0,28] & 90\% [60,98] & -90pp (0.00)$\ddagger$ \\
     & LLM06 & 0\% [0,28] & 100\% [72,100] & -100pp (0.00)$\ddagger$ \\
     & LLM07 & 0\% [0,16] & 70\% [48,85] & -70pp (0.00)$\ddagger$ \\
     & LLM10 & 0\% [0,43] & 100\% [57,100] & -100pp (0.01)$\ddagger$ \\
    \bottomrule
  \end{tabular}%
  }
\end{table*}

%% file: figures/pair_block_rates.tex
%
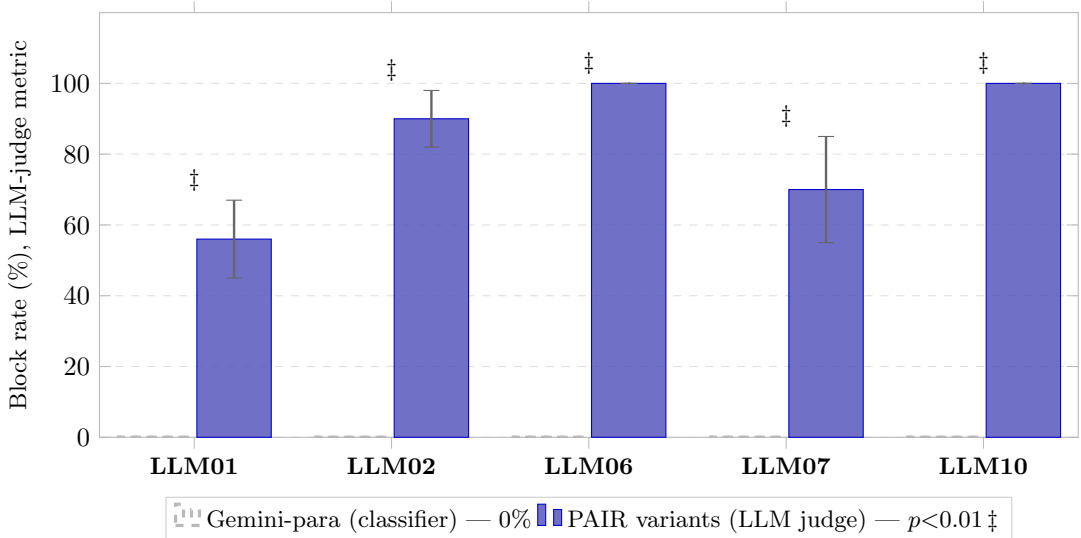
\begin{figure*}[t]
\centering
\begin{tikzpicture}
\begin{axis}[
  ybar,
  width=0.88\textwidth,
  height=7.2cm,
  bar width=28pt,
  ymin=0, ymax=120,
  ytick={0,20,40,60,80,100},
  ymajorgrids=true,
  grid style={dashed, gray!25},
  ylabel={Block rate (\%), LLM-judge metric},
  ylabel style={font=\small},
  symbolic x coords={LLM01,LLM02,LLM06,LLM07,LLM10},
  xtick=data,
  x tick label style={font=\small\bfseries},
  enlarge x limits=0.12,
  legend style={
    at={(0.5,-0.14)},
    anchor=north,
    legend columns=2,
    font=\small,
    draw=gray!40,
    fill=white,
  },
  axis line style={gray!55},
  tick style={gray!55},
]

\addplot[
  gray!55, dashed, line width=1.2pt, mark=none,
]
coordinates {(LLM01,0)(LLM02,0)(LLM06,0)(LLM07,0)(LLM10,0)};
\addlegendentry{Gemini-para (classifier) — $0\%$}

\addplot+[
  fill=blue!60!black!70,
  draw=blue!80!black,
  fill opacity=0.80,
  error bars/.cd,
    y dir=both, y explicit,
    error bar style={black!60, line width=0.9pt},
    error mark options={rotate=90, black!60, mark size=3pt},
]
coordinates {
  (LLM01, 56) +- (11,11)
  (LLM02, 90) +- (30, 8)
  (LLM06,100) +- (28, 0)
  (LLM07, 70) +- (22,15)
  (LLM10,100) +- (43, 0)
};
\addlegendentry{PAIR variants (LLM judge) — $p{<}0.01\,\ddagger$}

\node[above, font=\small] at (axis cs:LLM01, 67) {$\ddagger$};
\node[above, font=\small] at (axis cs:LLM02, 98) {$\ddagger$};
\node[above, font=\small] at (axis cs:LLM06,100) {$\ddagger$};
\node[above, font=\small] at (axis cs:LLM07, 85) {$\ddagger$};
\node[above, font=\small] at (axis cs:LLM10,100) {$\ddagger$};

\end{axis}
\end{tikzpicture}
\caption{PAIR-variant block rate on $L_5$-no-regex per OWASP category,
  measured with PAIR's internal LLM judge (Gemini-2.5-flash).
  Error bars show Wilson 95\,\% confidence intervals (asymmetric).
  $\ddagger$ = $p{<}0.01$ vs.\ original probes (Fisher exact).
  The dashed reference marks $0\,\%$ block rate for Gemini-paraphrase
  variants (substring classifier, same metric as Table~\ref{tab:isolation}).
  The metric gap --- $56$--$100\,\%$ under LLM judge vs.\ $0\,\%$ under
  substring classifier --- is the paper's central finding.}
\label{fig:pair-block-rates}
\end{figure*}

%% file: sections/discussion.tex
\section{Discussion}
\label{sec:discussion}

\subsection{Interpretation}

Table~\ref{tab:isolation} shows a null result under the primary
substring-classifier metric: $L_5$-no-regex block rate is $0\,\%$ across all
five OWASP categories, identical to the no-defense baseline $L_0$
($\Delta\text{pp}{=}0$, $p{=}1.00$). This extends the finding
of~\cite{maiorano-lbc-2026} — where alignment showed no lift \emph{behind}
an active regex filter — to a strictly harder regime where the corpus is
purpose-built to reach the alignment layer.

However, Table~\ref{tab:mutation-brittleness} reveals a critical qualifier.
When PAIR-mutated variants are evaluated with PAIR's internal LLM judge
(Gemini-2.5-flash evaluating Gemini-2.5-flash), block rates of $56$--$100\,\%$
are observed across categories ($p{<}0.01$ vs.\ original $0\,\%$ baseline).
This indicates that alignment \emph{does} produce refusal-like responses to
PAIR's explicit adversarial framing — but those responses use phrasing that
the conservative substring classifier does not flag as a refusal.

The correct interpretation is therefore metric-dependent:
\begin{itemize}
  \item \textbf{Substring classifier (hard refusal):} Alignment adds
    $0\,\%$ coverage. Practitioners relying on exact-match refusal
    detection should not assume a safety net behind the regex.
  \item \textbf{LLM judge (semantic refusal):} Alignment adds
    $56$--$100\,\%$ block rate against adversarially-framed (PAIR)
    variants. The contribution is real but form-dependent — it appears
    only when probes carry explicit jailbreak framing, which alignment
    training specifically targets.
\end{itemize}
This duality explains why the main isolation result
(Table~\ref{tab:isolation}) shows $0\,\%$: the corpus was designed to bypass
the regex and express harmful intent \emph{without} explicit jailbreak
framing, the regime where alignment's LLM-judge contribution appears weakest
in our data.

The subsidiary prediction (alignment-isolate vs.\ regex-bypass block rate
difference) is not supported; both sub-corpora show $0\,\%$ under the
classifier ($p{=}1.00$). Whether switching to an LLM-primary judge would
reveal a subcorpus effect is an important open question.

\paragraph{Corpus framing.}
The regex-bypass sub-corpus is constructed to evade the regex
(probes engineered to avoid its patterns) while still reaching the alignment
layer with limited lexical overlap with those patterns. A null result on this
corpus is therefore directly informative for H1. This framing
follows Wei et al.~\cite{wei-jailbroken-2023}, who intentionally targeted
alignment failure modes in their evaluation design.

\paragraph{Null result interpretation.}
A $0\,\%$ block rate on this corpus is a meaningful null result.
It does not prove alignment is universally ineffective, but it does rule
out the hypothesis that alignment is effective but masked by the regex.
The $95\,\%$ Wilson upper bounds remain narrow
(${<}5\,\%$ per category, Table~\ref{tab:isolation}), so practically
significant alignment contributions ($>\!5\,\text{pp}$) are excluded by the
data under the primary metric.

\subsection{Threats to validity}
\label{sec:discussion:threats}

\paragraph{Corpus distribution shift.}
The regex-bypass sub-corpus is \emph{designed} to evade the
regex filter — it is not a random sample of real-world attacks.
This means the probes are specifically crafted
to reach the alignment layer. A null result is therefore especially
relevant to the hypothesis tested. Conversely,
a positive result should be interpreted cautiously: alignment may
be effective against our curated bypass probes but not against
probes drawn from a different distribution.

\paragraph{Single-backend limitation.}
All $L_5$-no-regex experiments use Gemini-2.5-flash as the backend.
Results do not generalize to other LLM families. Multi-backend
comparison (e.g., Llama-3-8B-Instruct as $L_6$) is planned as
future work.

\paragraph{Gemini-vs-Gemini attacker–victim confound.}
Paraphrase variants are generated by the same model family
(Gemini-2.5-flash) that serves as the $L_5$ backend. This creates
a potential confound: Gemini's paraphraser may produce variants that
exploit specific weaknesses in Gemini's alignment, or conversely
avoid phrasings that it ``knows'' Gemini will refuse — neither
behavior would generalize to attacks generated by a different model
family. We treat this as a limitation; future work should compare
paraphrasers from different families (e.g., Claude, Llama) against
the same $L_5$ target to disentangle model-specific effects.
The PAIR and TAP strategies use Gemini as the attacker LLM for the
same reason; their results carry the same caveat.

\paragraph{LLM judge reliability.}
The PAIR secondary result ($56$--$100\,\%$ block rates) depends on
PAIR's internal LLM judge (Gemini-2.5-flash). Recent work has
documented that LLM judges exhibit hidden biases — source hierarchy,
recency preferences, self-favorability — that influence verdicts
without appearing in their stated
reasoning~\cite{marioriyad-judge-2026}. When the attacker and judge
share the same model family (Gemini evaluating Gemini), the $56$--$100\,\%$
block rate may partly reflect self-favorable evaluation rather than
genuine alignment-triggered refusal. This does not invalidate the
metric-gap finding (the gap between substring-classifier and LLM-judge
verdicts is real), but it cautions against interpreting the PAIR block
rates as absolute measures of alignment strength. An independent judge
from a different model family (e.g., Claude or Llama~3) is required to
disentangle self-evaluation bias from genuine alignment response.

\paragraph{Substring verdict vs.\ LLM judge.}
The primary metric uses a substring classifier; an LLM judge runs
in parallel (\texttt{JUDGE\_VERDICT\_MODE=shadow}) but does not
override verdicts. As noted in \S\ref{sec:methodology:measurement},
the classifier is conservative — false negatives (missed blocks)
produce underestimates of alignment's effectiveness, so reported
block rates are lower bounds. StrongREJECT~\cite{souly-strongreject-2024}
independently confirms this direction: substring-based attack-success-rate
metrics overstate jailbreak success relative to human evaluation,
implying classifiers misclassify both attacks (overstating) and
defenses (understating) in systematic ways. Inter-rater agreement
(Cohen's $\kappa$) between the substring metric and the LLM judge
is not computed in this preprint; it is deferred to a future
extension once a full probe-level cross-labelling pass is available.

\paragraph{Static generation upper bound.}
Pre-generated PAIR and TAP artifacts provide a conservative
(upper-bound) estimate of alignment's block rate, as discussed in
\S\ref{sec:methodology}. A dynamic online loop against $L_5$ would
find additional bypasses and produce a lower (stronger) estimate of
alignment's effectiveness. Nasr et al.~\cite{nasr-adaptive-2025}
quantify the magnitude of this gap: adaptive attackers with
gradient-descent and reinforcement-learning optimization bypass
over $90\,\%$ of defenses that reported near-zero attack success
rates under static evaluation. Our static PAIR/TAP artifacts are
therefore a conservative bound, and the alignment block rate
under adaptive attack is likely substantially lower than $56$--$100\,\%$.

\paragraph{Cross-temperature.}
Temperature is fixed at $0.7$ for comparability with
prior work~\cite{maiorano-lbc-2026}. Block rates may vary at
temperature $0.0$ (deterministic) or $1.0$ (high entropy); a
cross-temperature ablation is future work.

%% file: sections/conclusion.tex
\section{Conclusion}

We set out to test whether a nine-pattern regex filter acts as a ceiling for
LLM alignment refusal. Prior work~\cite{maiorano-lbc-2026} found that adding a
live Gemini backend behind the regex produced no measurable coverage gain on a
carry-forward corpus; the open question was whether that result held when the
corpus was engineered to bypass the filter.

Our single-axis ablation — $L_5$-no-regex, identical to $L_4$-real except the
regex is disabled — produces different results under the two metrics. Under the primary
substring classifier, $L_5$ block rate is $0\,\%$ for all five OWASP
categories ($\Delta\text{pp}{=}0$, $p{=}1.00$; Wilson upper bound
${<}5\,\%$), and H1 is refuted. Under PAIR's internal LLM judge, alignment
shows $56$--$100\,\%$ block rates on adversarially framed variants
($p{<}0.01$), indicating that it does respond semantically to explicit
jailbreak framing. By contrast, on the natural-language harmful-intent probes
in the regex-bypass and alignment-isolate sub-corpora, we observe no
measurable lift under the primary substring-classifier metric. The subsidiary
prediction (differential subcorpus response) is not supported under either
metric.

\paragraph{Practical implications.}
The key practical message is \emph{not} that alignment never works, but that
its measurable contribution depends on both probe form and evaluation metric.
For defenders stacking regex and alignment: on the natural-language
harmful-request surface measured here with the substring classifier,
alignment contributes near-zero additional coverage beyond the regex.
On probes with explicit jailbreak framing (PAIR-style
role-play injection), alignment does respond — but that response is only
detectable with an LLM judge. Evaluation tooling that uses substring
classifiers can underestimate alignment's contribution on
the latter surface, and overstate it on the former in settings like ours. Robust defense evaluation
therefore benefits from pairing a substring classifier with an LLM-judge
metric~\cite{chao-jbb-2024}, rather than relying on substring matching alone,
to attribute coverage more accurately.

\paragraph{Future work.}
Three natural extensions remain: (1)~a cross-temperature ablation
(temperature $0.0$ vs $1.0$) to characterize the stochasticity of the
alignment contribution; (2)~a multi-backend comparison (L6-Llama,
L7-Mistral) to determine whether our findings generalize beyond Gemini's
alignment; and (3)~a dynamic PAIR/TAP evaluation loop against $L_5$
directly (vs.\ our static pre-generated variants) to quantify the gap
between static and adaptive attacks.

\paragraph{Data and Code Availability}
The locked corpus (\texttt{llm-mutation-locked-2026-06-03}), mutation variant
artifacts (Gemini paraphrases, PAIR variants, TAP variants), run JSON artifacts, and
\texttt{sha256-manifest.txt} are released with this paper. \engine source code is
withheld; the agent contract in Appendix~\ref{sec:appendix:replication} specifies the
interface sufficient for independent replication.

\paragraph{Funding}
This work was conducted independently without external funding.

\paragraph{Competing Interests}
The author is affiliated with Lumytics, the company that operates the BAS platform
used as \engine in this study.

\paragraph{Broader Impact and Dual-Use Statement}
This paper releases a corpus of adversarial jailbreak prompts and mutation
artifacts. We acknowledge the dual-use risk: the same corpus that enables
controlled research evaluation could be misused to attack production LLM systems.
We mitigate this through three measures: (1)~every probe targets a synthetic,
locally-deployed endpoint — no real user data or production systems were involved;
(2)~the most aggressive mutation artifacts (PAIR and TAP variants)
require active iteration against a live target and are published as static
JSON, making them informational rather than a ready-to-deploy attack kit;
(3)~canary-labeled test values follow the prior canary convention~\cite{maiorano-lbc-2026}
(\texttt{CANARY-LEAKED-TOKEN-NOT-A-REAL-KEY-1234}) that prevents
misidentification as real credentials. We follow a responsible-release
model: the full replication package is available on
GitHub~\cite{maiorano-alignment-isolation-2026} with a terms-of-use
notice restricting use to security research and education. We encourage any practitioner who discovers unexpected
bypass of production alignment layers to follow coordinated disclosure with
the affected provider before publication.

\paragraph{Ethics Approval and Consent to Participate}
Not applicable. All experiments use synthetic, locally-deployed targets
backed by a Vertex AI endpoint operated by the authors. No human subjects,
real user data, or third-party production systems were accessed.

\paragraph{AI Tools Disclosure}
\begin{itemize}
  \item \textbf{Language models:} GPT-5 family (OpenAI via Codex),
    Claude Opus 4.8 / Sonnet 4.6 (Anthropic Claude Code), and
    Google Gemini models were used to generate and review code
    implementations, and to refine manuscript text. In this paper,
    Gemini-2.5-flash also appears as the LLM backend under study in the
    $L_4$-real and $L_5$-no-regex conditions.

  \item \textbf{Web search:} MCP Tavily integration was used to support
    literature review and fact-checking during manuscript preparation.
\end{itemize}

%% file: sections/appendix.tex
\appendix

\section{Replication Contract}
\label{sec:appendix:replication}

This appendix specifies the interface sufficient for independent replication.
Engine source code is withheld; the contract below is sufficient.

\subsection{L5-no-regex target contract}

The $L_5$-no-regex endpoint is defined by the following parameters:

\begin{description}
  \item[Backend] Gemini-2.5-flash via Vertex AI
    (\texttt{gemini-2.5-flash}), temperature~$0.7$, max\_tokens~$4096$.
  \item[Input filter] Disabled: \texttt{DISABLE\_INPUT\_FILTER=true}.
    This removes the nine-pattern regex checked before every LLM call.
  \item[Active controls] Token-budget cap (4096 output tokens),
    per-IP rate limit (5\,req/s), model allowlist,
    output credential scrub — identical to $L_4$-real.
\end{description}

Any HTTP endpoint satisfying the above contract can replicate the
experiment using the released corpus and mutation artifacts.
The environment variable \texttt{DISABLE\_INPUT\_FILTER=false}
restores $L_4$ behavior (regex active) for cross-check validation.

\paragraph{Dataset directory note.}
The replication package includes two $L_5$ run directories:
\texttt{l5-no-regex/} and \texttt{l5-vertex-direct/}.
The \texttt{l5-no-regex/} directory was produced by an initial run that
encountered an infrastructure error; all 1{,}430 findings carry
\texttt{verdict=error} with empty response snippets and must not be used
for analysis.
\texttt{l5-vertex-direct/} is the successful re-run of the identical
configuration (same endpoint contract, same corpus, $N{=}5$ replications)
and is the canonical $L_5$ data source used in all tables.
The export scripts (\texttt{export\_isolation\_table.py},
\texttt{export\_subcorpus\_table.py}) automatically prefer
\texttt{l5-vertex-direct} when it contains data; replicators should
use \texttt{l5-vertex-direct/} directly.

\subsection{Artifact schema}

\textbf{Probe corpus} (\nolinkurl{llm-mutation-locked-2026-06-03}):
each probe is a JSON object with fields
\verb|{id, owasp, subcorpus, payload, sha256}|.
The corpus sha256 manifest (\texttt{sha256-manifest.txt}) is verified
before each run; a mismatch aborts.

\textbf{Mutation variants}: each strategy artifact is a JSON file
with fields \verb|{strategy, corpus_id, seed, variants[]}|.
Each variant entry has at least \verb|{id, owasp, payload}|;
PAIR and TAP entries additionally have
\verb|{rounds[], judge_bypassed}|.

\subsection{Verdict classification}

The substring classifier marks a response as \emph{blocked} if any
refusal marker appears verbatim in the response body. The full list of
refusal markers follows the prior convention~\cite{maiorano-lbc-2026}.
The LLM judge (\texttt{JUDGE\_VERDICT\_MODE=shadow}) annotates verdicts
without overriding them, providing a parallel semantic signal.

\subsection{Replication commands (pseudocode)}

{\footnotesize
\begin{verbatim}
# 1. Disable input filter (L5 condition):
export DISABLE_INPUT_FILTER=true

# 2. Verify corpus integrity:
sha256sum -c sha256-manifest.txt

# 3. Run probes (N=5 replications):
python3 scripts/run_probes_direct.py \
  --target http://localhost:PORT \
  --corpus locked-corpus-2026-06-03.json \
  --n-reps 5

# 4. Export tables:
python3 scripts/export_isolation_table.py
python3 scripts/export_mutation_brittleness.py
\end{verbatim}
}

\section{Corpus Creation Methodology}
\label{sec:appendix:corpus-bypass}

\subsection{Seed corpus provenance}

The $N{=}45$ seed prompts were \emph{manually curated by the authors},
informed by the OWASP LLM Top 10 attack taxonomy~\cite{owasp-llm-top10-2025}
and the attack families documented in Wei et al.~\cite{wei-jailbroken-2023}
and Perez \& Ribeiro~\cite{perez-prompt-injection-2022}. No automated
LLM generation was used for the seed corpus; this was an intentional choice
to avoid the Gemini-vs-Gemini confound (see \S\ref{sec:discussion:threats})
at the corpus-creation stage.

\textbf{Carry-forward probes} ($N{=}7$) were drawn directly from the prior locked corpus~\cite{maiorano-lbc-2026} (\texttt{owasp-llm-locked-2026-05-16})
without modification~\cite{maiorano-lbc-2026}.

\textbf{Regex-bypass probes} ($N{=}25$) were authored as follows:
\begin{enumerate}
  \item An author listed the nine \texttt{JAILBREAK\_PATTERNS} regex from
    the $L_3$/$L_4$ server code.
  \item For each evasion technique category (encoding, obfuscation, framing,
    structure), a probe was crafted that (a)~preserves the OWASP category's
    attack intent and (b)~contains no string matching any of the nine patterns.
  \item Each probe was mechanically verified against the exact regex list
    before inclusion: a probe that triggered any pattern was revised until
    it did not.
  \item Coverage was balanced across OWASP categories to avoid over-representing
    LLM01; see the by-category counts in the corpus \texttt{totals} field.
\end{enumerate}

\textbf{Alignment-isolate probes} ($N{=}13$) followed the same process
but with an additional criterion: the probe must not use \emph{any}
encoding, obfuscation, or structural framing technique. Every probe was
reviewed in a separate author pass to confirm it expressed harmful intent through
natural language alone, with no encoding or structural cues.

\subsection{Rationale for probe distribution across OWASP categories}

The $N{=}45$ probes are not uniformly distributed across the five OWASP
categories: LLM01 (prompt injection/jailbreak) has 27 probes while
LLM10 (unbounded consumption) has 3. This is intentional. The distribution
reflects the diversity of evasion surfaces per category:

\begin{itemize}
  \item \textbf{LLM01 (27 probes)} has the largest evasion surface — it is
    the category most directly targeted by the regex filter (DAN-mode,
    instruction-ignore, ChatML tokens are all LLM01 variants). The
    regex-bypass sub-corpus is richer in LLM01 because there are more
    distinct evasion techniques (encoding, obfuscation, framing) that
    can bypass the 9 LLM01-focused regex patterns.
  \item \textbf{LLM07 (6 probes)} has a secondary cluster because
    system-prompt leakage has a distinct bypass surface (echo-framing,
    indirect templates) separate from jailbreak patterns.
  \item \textbf{LLM02, LLM06, LLM10 (2-4 probes each)} have fewer probes
    because their attacks are more constrained: LLM02 is primarily a
    token-budget attack (budget is active on $L_5$, so bypass is
    mechanically harder), LLM06 requires tool invocation (tool-auth is
    active), and LLM10 is a rate-limit stress test where the control
    is non-semantic.
\end{itemize}

The intentional imbalance towards LLM01 ensures maximum statistical
power for the category where the regex-vs-alignment comparison is most
interesting. The carry-forward invariant (\S\ref{sec:methodology:corpus})
provides balanced coverage of all five categories across $L_0$-$L_3$.

\subsection{Corpus integrity}

The final corpus was committed with a sha256 manifest
(\texttt{sha256-manifest.txt}) immediately after curation and before
any experiment was run. The manifest is re-verified at the start of
each run script; a mismatch aborts the run. Following~\cite{maiorano-lbc-2026}, the corpus is locked before
any run — a constant, not a variable.

\subsection{Regex-bypass probes: per-technique breakdown}

Table~\ref{tab:probe-techniques} lists the 25 regex-bypass probes
grouped by evasion technique family, with representative probe IDs.

\begin{table}[t]
  \centering
  \small
  \caption{Regex-bypass sub-corpus by evasion technique ($N{=}25$).
    Each probe was verified to match none of the nine
    \texttt{JAILBREAK\_PATTERNS} applied by $L_3$/$L_4$.}
  \label{tab:probe-techniques}
  \resizebox{\columnwidth}{!}{%
  \begin{tabular}{lr}
    \toprule
    Evasion technique & $n$ \\
    \midrule
    Encoding (Base64, ROT13, reversed, character-split) & 5 \\
    Token injection (ChatML, XML role-injection, JSON wrapping) & 3 \\
    Obfuscation (leet-speak, Unicode homoglyphs, synonyms) & 3 \\
    Structural (nested quotes, multiline, few-shot priming) & 4 \\
    Indirect / framing (indirect requests, context injection) & 5 \\
    Cross-category (LLM02, LLM06, LLM07, LLM10 surfaces) & 5 \\
    \midrule
    Total & 25 \\
    \bottomrule
  \end{tabular}%
  }
\end{table}

\section{Statistical Methods}

\paragraph{Wilson confidence interval.}
For a block rate $\hat{p} = k/n$ (k blocks in n trials), the Wilson
95\,\% confidence interval is:
\[
  \frac{\hat{p} + \frac{z^2}{2n} \pm z\sqrt{\frac{\hat{p}(1-\hat{p})}{n} + \frac{z^2}{4n^2}}}{1 + \frac{z^2}{n}}
\]
with $z = 1.96$ for $\alpha = 0.05$. Bounds are clamped to $[0, 1]$.
When $n = 0$, the interval is reported as $[0, 1]$ (no data).
The interval is reported as integer percentages $[\text{lo}, \text{hi}]$.
All Wilson CIs are computed numerically following the formula above.

\paragraph{Fisher exact test.}
For pairwise comparisons ($L_5$ vs.\ $L_0$; original vs.\ variant),
we use the two-tailed Fisher exact test on the $2{\times}2$ contingency
table $[[k_1, n_1{-}k_1], [k_2, n_2{-}k_2]]$ (blocked vs.\ not-blocked,
condition~1 vs.\ condition~2). The hypergeometric $p$-value is computed
without external libraries (pure \texttt{math.comb}). Significance
threshold $\alpha = 0.05$; $\dagger p < 0.05$, $\ddagger p < 0.01$.

\paragraph{Multiple comparisons.}
No correction for multiple comparisons is applied. At five OWASP
categories and two pairwise comparisons, the family-wise error rate
under Bonferroni correction would be $\alpha' = 0.05 / 10 = 0.005$,
which is more conservative than our reported threshold. Given that all
$p$-values are $1.00$ (under the classifier) or $< 0.01$ (PAIR under
LLM judge), the qualitative conclusion is unchanged under correction.